\documentclass{llncs}
\usepackage{amsmath}

\usepackage{graphicx}
\usepackage{mygraphics}

\begin{document}

\title{Turkish PoS Tagging by Reducing Sparsity with Morpheme Tags in Small Datasets}
\titlerunning{Turkish PoS Tagging by Reducing Sparsity with Morpheme Tags in Small Datasets}

\author{Burcu Can\inst{1} \and Ahmet \"{U}st\"{u}n\inst{2} \and Murathan Kurfal{\i}\inst{2}}
\authorrunning{Burcu Can et al.} 
%
\tocauthor{Burcu Can, Ahmet \"{U}st\"{u}n, Murathan Kurfal{\i}}
\institute{Department of Computer Engineering, Hacettepe University \\ 
Beytepe, Ankara, 06800, Turkey\\
\email{burcucan@cs.hacettepe.edu.tr}
\and
Cognitive Science Department, Informatics Institute \\ Middle East Technical University (ODT\"U) \\
Ankara, 06800, Turkey \\
\email{\{ustun.ahmet,kurfali\}@metu.edu.tr}
}

\maketitle

\begin{abstract}
Sparsity is one of the major problems in natural language processing. The problem becomes even more severe in agglutinating languages that are highly prone to be inflected. We deal with sparsity in Turkish by adopting morphological features for part-of-speech tagging. We learn inflectional and derivational morpheme tags in Turkish by using conditional random fields (CRF) and we employ the morpheme tags in part-of-speech (PoS) tagging by using hidden Markov models (HMMs) to mitigate sparsity. Results show that using morpheme tags in PoS tagging helps alleviate the sparsity in emission probabilities. Our model outperforms other hidden Markov model based PoS tagging models for small training datasets in Turkish. We obtain an accuracy of 94.1\% in morpheme tagging and 89.2\% in PoS tagging on a 5K training dataset.

\end{abstract}
\section{Introduction}
Turkish is an agglutinating language that builds words by gluing meaning bearing units called morphemes. While gluing morphemes together, vowel harmony and consonant assimilation are intensely applied leading to orthographic transformations in morphemes. For example, the suffix \textit{dir} can be transformed into \textit{d{\i}r}, \textit{dur}, \textit{d\"{u}r} depending on the last vowel in the word to which it is being attached. This is called vowel harmony. Moreover, the same morpheme can be transformed into \textit{tir}, \textit{t{\i}r}, \textit{tur}, \textit{t\"{u}r}, this time depending on the last consonant of the word. This is called consonant assimilation. Both vowel harmony and consonant assimilation introduce different realizations of the same morpheme, which are called allomorphs (e.g. \textit{dir}, \textit{d{\i}r}, \textit{dur}, \textit{d\"{u}r}, \textit{tir}, \textit{t{\i}r}, \textit{tur}, \textit{t\"{u}r} are all allomorphs). 

Agglutination already introduces a sparsity problem in natural language processing for especially agglutinating languages. The sparsity problem becomes more crucial when a morpheme has got different realizations. Identifying morphemes that are realizations of each other is the starting point of this work.

Morphological segmentation systems normally provide only the segments of words without any morpheme tags. However, labeled segmentation is required for some natural language processing tasks. For example, in sentiment analysis the Turkish negation suffix \textit{ma} (and its allomorph \textit{me}) needs to be distinguished from the derivational suffix \textit{ma} (and its allomorph \textit{me}) that turns a verb into a noun in order to extract the correct sentiment out. The same also applies for machine translation, question answering, and other natural language processing applications. 

Morpheme tagging has become a neglected aspect of morphological segmentation. In this paper, we use conditional random fields (CRF) for morpheme tagging in a weakly-supervised setting. We use the obtained morpheme tags in part-of-speech tagging (PoS tagging) in order to mitigate sparsity in a case when small amount of data is provided. Indeed the sparsity problem is quite severe in PoS tagging for especially agglutinating languages where different methods (e.g. smoothing) have been applied to deal with sparsity. The sparsity is alleviated significantly by using morpheme tags rather than using lexical instances such as words or suffixes. 

This paper is organized as follows: Section~\ref{relatedwork} points at the related work in the literature, section~\ref{model} describes the CRF model adopted in morpheme tagging and describes the HMMs used in PoS tagging, section~\ref{experiments} presents the experimental results from both tasks and finally section~\ref{conclusion} concludes the paper with the remaining future work.

\section{Related Work}
\label{relatedwork}
There has been a substantial amount of work on unsupervised morphological segmentation. Goldsmith~\cite{Goldsmith}, Creutz and Lagus~\cite{Creutz2002} build morphological segmentation systems based on minimum description length (MDL). Creutz and Lagus~\cite{CreutzCatMap} introduce a hidden Markov model (HMM) that employs the probability distributions between different morpheme categories such as prefix, stem, and suffix. Poon et al.~\cite{Poon} introduce a log-linear model for unsupervised morphological segmentation that incorporates MDL-inspired priors. 

All of these models provide only morphological segmentations of words and not any morphological tags that identify the morpheme roles within a word. Learning morpheme tags involve distinguishing homophonous morphemes\footnote{Morphemes with the same surface forms but with different meanings.} and learning allomorphs. Oflazer~\cite{Oflazer93} introduce derivational boundaries and inflectional groups in Turkish morphological analysis. This is performed by two-level morphology (PC-KIMMO~\cite{Antworth,Kimmo1983}) that formulates morphological segmentation via a cascade of finite state transducers by employing morphophonemic alternations. All ortographic and morphophonemic rules are implemented by a set of finite-state automata (FSA) rules. Their model gives a labeled morphological analysis based on these rules. 

Allomorfessor~\cite{Allomorfessor} is one of the models that aims to perform morphological segmentation based on allomorphs by modeling mutations between different surface forms of morphemes, namely allomorphs. Can and Manandhar~\cite{canmanandhar} develop an agglomerative hierarchical clustering to find the morpheme classes in an unsupervised setting. 

To our knowledge, Ryan et al.~\cite{Ryan} introduce labeled morphological segmentation for the first time in a supervised learning framework without using any rules. They model morphotactics by a semi-Markov model. Different levels of tagsets are introduced that capture different levels of granularity. Our model resembles their model from the aspect of morphological tagging. 

Morpheme tags have been used in many natural language processing tasks. El-Kahlout and Oflazer~\cite{elkahlout} employ morphological tags in order to alleviate the sparsity by matching the Turkish morphemes having the same morphological tag to the same English translation in statistical machine translation task. They address that using morphological tags provides a substantial improvement on the BLUE score. 

Morpheme tags have been used in morphological/PoS disambiguation in Turkish language. Ehsani et al.~\cite{ehsani} use conditional random fields for disambiguating PoS tags in Turkish by utilizing the morphological tags. They introduce some dependencies between inflectional groups of morphemes in order to simplify the transition probabilities. Sak et al.~\cite{Sak2007} apply perceptron algorithm for morphological disambiguation. Hakkani-Tur et al.~\cite{hakkanitur} formulate a trigram HMM based on inflectional groups in order to disambiguate morphological parses of a given word. The results show that using the dependencies between inflectional groups of adjacent words improve PoS tagging accuracy. Many of these models select a complete morphological analysis for each word rather than providing a single PoS tag. 

Dincer et al.~\cite{dincer} formulate HMMs by emitting suffixes rather than emitting words in order to mitigate the sparsity. However, they do not use any morpheme tags. Our PoS model is mostly similar to their work in this respect. We use morpheme tags in order to cope with the sparsity in emission probabilities rather than using fixed-length endings of words. 

\section{Model}
\label{model}

\subsection{Turkish Morphology}

Turkish is an agglutinating language that has a productive inflectional and derivational suffixation. This brings the sparsity problem in nlp tasks due to the large vocabulary introduced by the language. The vocabulary size of a corpus having 1 million words becomes 106.547~\cite{hakkanitur}. In order to deal with the sparsity, a representation that shows inflectional groups and derivation boundaries of the morphological analysis of each word is introduced by Hakkani-T\"{u}r et al.~\cite{hakkanitur}.The different morphological analyses of the word \textit{al{\i}nd{\i}} are given as follows by a two-level morphological analyzer~\cite{ehsani}: \\

  al+Verb\textasciicircum{}DB+Verb+Pass+Pos+Past+A3sg (\textit{it was taken})
 
 al+Adj\textasciicircum{}DB+Noun+Zero+A3sg+P2sg+Nom\textasciicircum{}DB+Verb+Zero+Past+A3sg (\textit{it was your red})
 
 al+Adj\textasciicircum{}DB+Noun+Zero+A3sg+Pnon+Gen\textasciicircum{}DB+Verb+Zero+Past+A3sg (\textit{it was the one of the red})
 
 al{\i}nd{\i}+Noun+A3sg+Pnon+Nom (\textit{receipt})
 
 al{\i}n+Verb+Pos+Past+A3sg (\textit{resent})
 
 al{\i}n+Noun+A3sg+Pnon+Nom\textasciicircum{}DB+Verb+Zero+Past+A3sg (\textit{it was the forehead})  \\
 
Here ˆDB's denote the derivation boundaries and the rest of the morpheme tags denote the inflectional groups (IGs). Most of the words have more than one morphological analysis in Turkish and the morphological disambiguation aim to find the right morphological analysis of the word given in a specific context. 

In this work, we are only using the morpheme tags (both derivational and inflectional) of words in order to find a single PoS tag for each word. We believe that morpheme tags give the best clue for a PoS tag. This is sufficient if we are only interested in syntax but not in the meaning. For example, the analyses of \textit{al{\i}nd{\i}} that end with \textit{A3sg} can be considered as verbs, whereas the only analysis ending with \textit{Nom} can be considered as a noun. In order to find the morpheme tags we only use the morphotactic features of morphemes within the words, whereas we use contextual features and morphological features in PoS tagging. 

\subsection{Morphological Tagging by Using CRFs}

Conditional Random Fields (CRF)~\cite{lafferty2001conditional} are undirected graphical models that are generally used for segmenting and labeling a given sequence. Unlike HMMs, CRFs are discriminative models that define the conditional distribution \(P(Y|M) \) rather than the joint probability distribution \(P(M,Y) \), where Y corresponds to the label sequence $Y=\{y_0, y_1, \cdots, y_n\}$ and $M$ corresponds to the input (i.e. observations) sequence $M=\{m_0, m_1, \cdots, m_n\}$. In our case, the label sequence $Y$ refers to the morpheme tags and observation sequence $M$ refers to the morphemes.

The conditional distribution \(P(Y|M) \) in our CRF model is given as follows:

\begin{equation}
p(Y|M) = \frac{1}{Z(M)} \prod_n^N   \prod_i^{I_n}  \lambda F (Y,M)
\end{equation}
that iterates over the morphemes of each word in the corpus with $N$ words, each having $I_n$ morphemes defined on a feature set $F$. Here \({Z(M)}\)  is the normalization factor:

\begin{equation}
Z(M)= \sum_{y_i} \prod_n^N   \prod_i^{I_n}  \lambda F (Y,M)
\end{equation}
Here  \(\lambda\) corresponds to the weight vector for the feature set \(F\). Feature function $F$ consists of two types: state feature functions \(s(y,m,i)\) and transition feature functions \(s({y}',y,m,i)\) where $i$ denotes the input position. State feature function is non-zero when the label \(y_{i}\) is matched with the label defined in the function, whereas transition functions depend on the label sequence \(y_{i-1},y_{i}\).

\newsavebox{\crf}
\begin{lrbox}{\crf}
\cropgraphics[0.50]{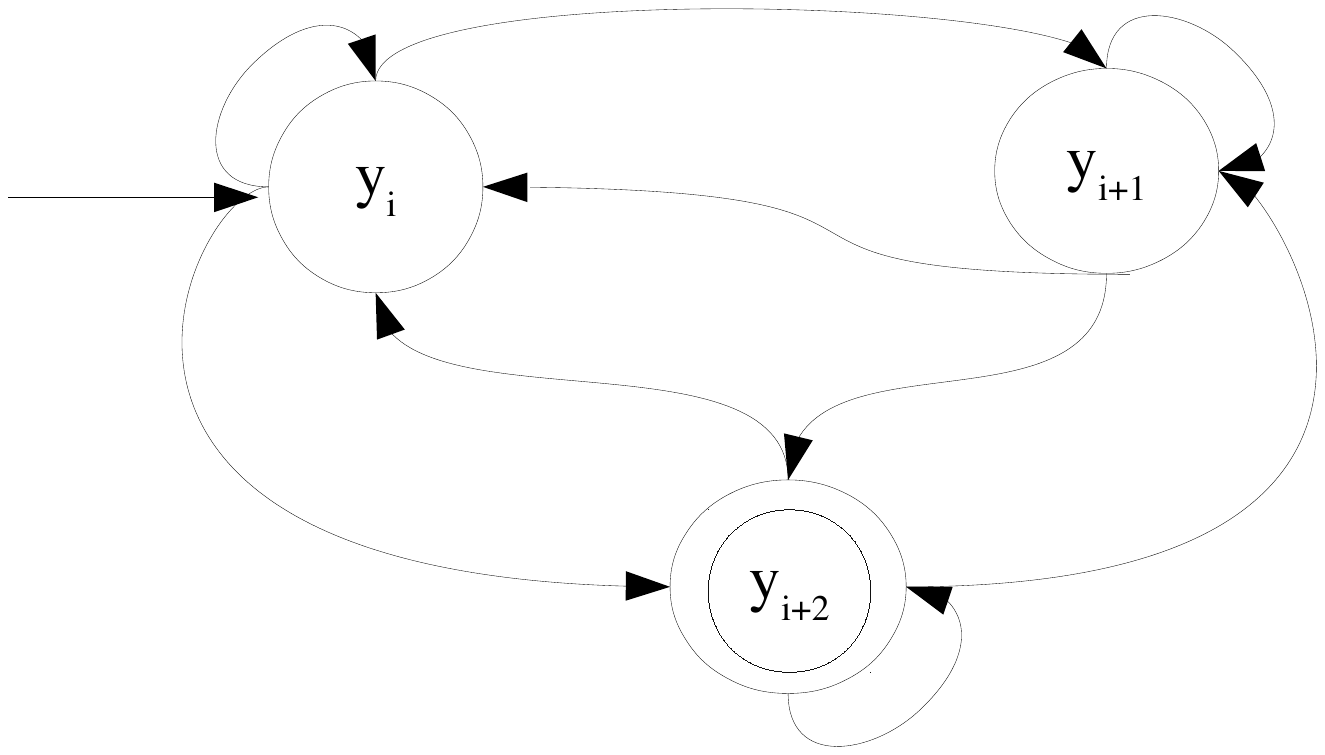}{100}{298}
\end{lrbox}

\begin{figure}[t!]
\begin{center}
\usebox{\crf}
\end{center}
\caption{Our Naive model for the conditional random field. Each state denotes a morpheme tag, where a word $w$ is defined as $w=m_i/y_{i}+m_{i+1}/y_{i+1}+m_{i+2}/y_{i+2}$.}\label{fig:crf}
\end{figure}

Our model is given in Figure~\ref{fig:crf}. We adopt a Naive model where an edge is built between every state pair. Therefore, morpheme tags within the same word are assumed to be dependent on each other, whereas each word is assumed to be independent from the others. Thus, we deal with only morphotactic rules within the same word for morpheme tagging task without using any contextual features. 

\subsection{Adopting Morphological Tags in PoS Tagging}

We use the obtained morpheme tags from the CRF model in order to infer the PoS tags of  words. We learn PoS tags according to the following formulation by finding the PoS tag sequence that maximizes the probability for a given sequence of words:

\begin{equation}
\label{posterior}
argmax_{t_{1}\cdots t_{n}} P(t_{1}\cdots t_{n}|w_{1}\cdots w_{n})
\end{equation}
where \(t_{1}\cdots t_{n}\) denotes the PoS tags and \(w_{1}...w_{n}\) denotes the sequence of words. The Bayes' rule is simply applied for the posterior probability as follows:

\begin{eqnarray}
\label{hmmformula}
\arg\max\limits_{t_{1}\cdots t_{n}} P(t_{1}\cdots t_{n}|w_{1}...w_{n}) &=& \arg\max\limits_{t_{1}\cdots t_{n}} \frac{P(w_{1}...w_{n}|t_{1}\cdots t_{n}) P(t_{1}\cdots t_{n})}{P(w_{1}\cdots w_{n})} \nonumber \\ 
         &\propto& \arg\max\limits_{t_{1}\cdots t_{n}} P(w_{1}\cdots w_{n}|t_{1}\cdots t_{n}) P(t_{1}\cdots t_{n})  \nonumber \\ 
\end{eqnarray}
where $P(w_{1}\cdots w_{n})$ is discarded since it is the same for all tag assignments for the given word sequence. 

We formulate the posterior probability as a trigram HMM by assuming that each PoS tag depends only on the previous two tags:

\begin{equation}
P(t_{1}\cdots t_{n}) = p(t_1) p(t_2|t_1)\prod_{i}p(t_i|t_{i-2},t_{i-1})
\end{equation}

We apply interpolation to smooth the transition probabilities in order to rule out zeros in transitions with the equation given below:

\begin{equation}
P_{inter}(t_{i}|t^{i-1}_{i-n+1}) = \beta_{t^{i-1}_{i-n+1}}P(t_{i}|t^{i-1}_{i-n+1})+(1-\beta_{t^{i-1}_{i-n+1}})P_{inter}(t_{i}|t^{i-1}_{i-n+2})
\end{equation}
which defines an nth-order smoothed model where \(P_{inter}(t_{i}|t^{i-1}_{i-n+1})\) corresponds to the transition probability after interpolation is applied recursively. We estimate the parameters $\mathbf{\beta}$ by tuning our model on a development set. 

\newsavebox{\hmm}
\begin{lrbox}{\hmm}
\cropgraphics[0.35]{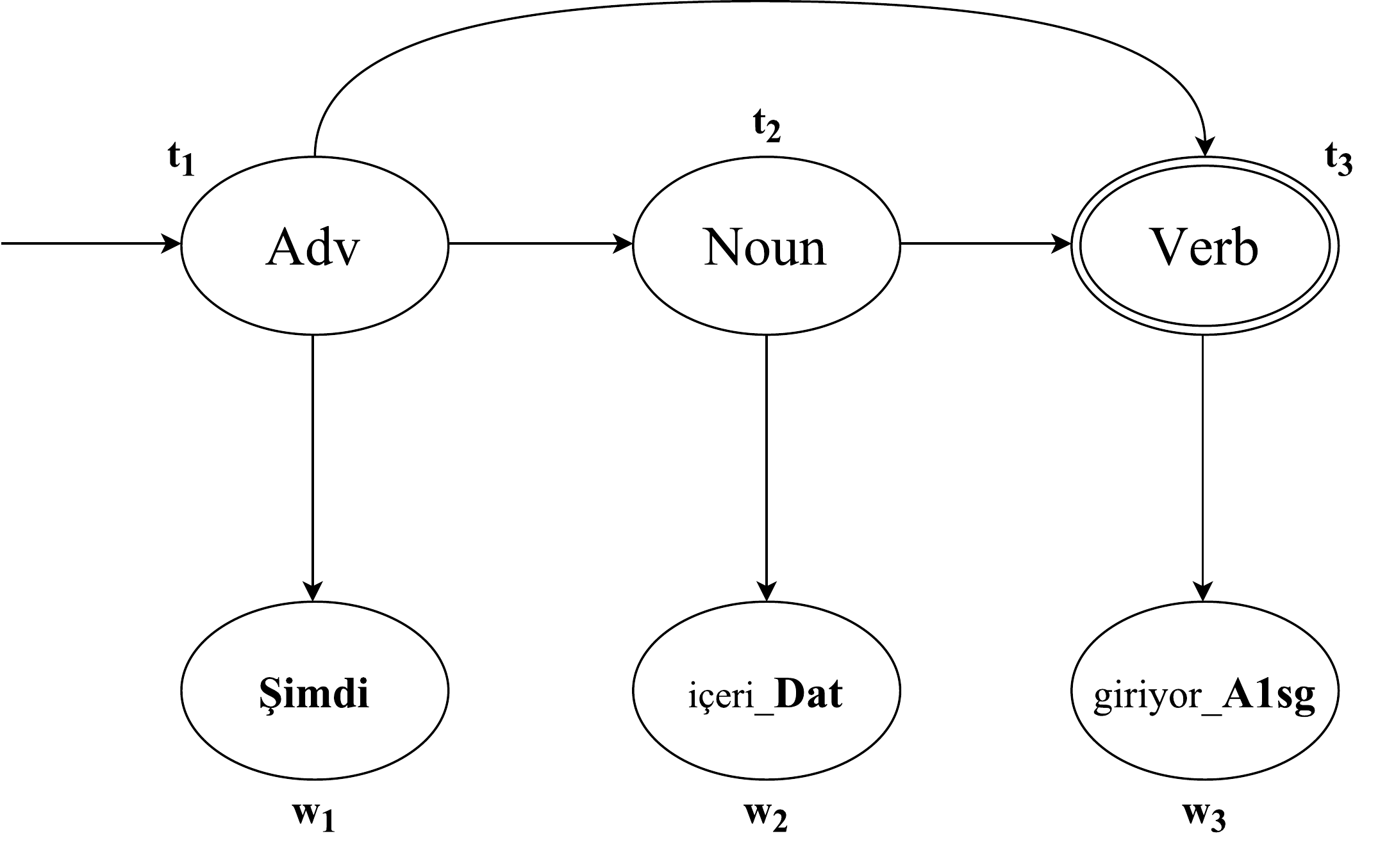}{10}{0}
\end{lrbox}

\begin{figure}[t!]
\begin{center}
\usebox{\hmm}
\end{center}
\caption{Our trigram HMM adopted for PoS tagging. The bold units are emitted from the states. The first word \textit{\c{S}imdi} is emitted from $t_1$, the tag of the last morpheme in the second word \textit{Dat} is emitted from $t_2$, and the tag of the last morpheme in the third word \textit{A1sg} is emitted from $t_3$.}\label{fig:hmm}
\end{figure}

The sparsity problem also emerges in the emission probabilities. We emit the tag of the last morpheme in the word if the word has more than two segments. Otherwise, the word itself is emitted from the PoS tag as seen in Figure~\ref{fig:hmm}. Therefore, the emission probabilities are estimated as follows:

\begin{equation}
p(w_i|t_i)=
\begin{cases}
    p(y_{n-1}|t_i),   & \text{if } w_i=\{m_1/y_1+\cdots +m_{n-1}/y_{n-1}\} \\
    p(w_i|t_i),& \text{otherwise}
\end{cases}
\end{equation}
where $y_{n-1}$ is the morpheme tag of the last suffix in the word. We apply interpolation to smooth $p(w_i|t_i)$ for the words which do not exist in the corpus and cannot be segmented further:

\begin{equation}
P_{inter}(w_{i}|t_{i}) = \alpha P(w_{i}|t_{i})+(1-\alpha)\frac{max(f(w_{i}),1)}{N}
\end{equation}
Here \(P_{inter}(w_{i}|t_{i})\) corresponds to the smoothed emission probabilities, \(f(w_{i})\) is the number of word tokens of type \(w_{i}\), \(N\) is the vocabulary size, and $\alpha$ is the interpolation coefficient.

Viterbi algorithm is applied to find the PoS tag sequence that maximizes the posterior probability given in Equation~\ref{hmmformula}.

\section{Experiments \& Results}
\label{experiments}

\subsection{Data}

We used several different corpora for the experiments. One of them is METU-Sabanc{\i} Turkish Treebank~\cite{say2002development} that consists of 56k word tokens and 5600 sentences. The dataset includes PoS tags and morphological analyses of the words. 

For the additional experiments, in order to compare our CRF model with the Semi-Markov Model by Ryan et al.~\cite{Ryan}, we used their dataset that consists of 3573 morphologically segmented and tagged word tokens, of which 1987 words belong to the train set and 1586 words belong to the test set. 

In order to compare our PoS tagging model with Sak et. al.~\cite{Sak2007}, we used their training and test set that are collected from various newspaper archives. This dataset consists of $\sim$800k word tokens and $\sim$47.5k sentences.

For all of the experiments, we used a separate development set that consists of 6K words to tune the interpolation coefficients. We assigned $\alpha=0.9$ for the emission probabilities, $\beta_1=0.6$ (bigram) and $\beta_2=0.4$ (unigram) for the bigram transition probabilities, and $\beta_1=0.5$ (trigram), $\beta_2=0.3$ (bigram), and $\beta_3=0.2$ (unigram) for the interpolation used in trigram transitions. 

For morpheme tagging experiments, we removed out all punctuation from the datasets and reintroduced the terminal punctuation for PoS tagging task since the word boundaries are crucial for PoS tagging.

\subsection{Experiments on Morphological Tagging}

In morpheme tagging task, we assume that morphological segmentations of words are provided. We obtained the segmentations and morpheme labels through an open-source morphological analyzer called Zemberek~\cite{akin2007zemberek} in order to build a train set for the morpheme tagging task. Zemberek defines 84 different morpheme tags on Metu-Sabanc{\i} Treebank. We used the open source CRF package~\cite{sha2003shallow} for training our own model on our training set. Some of the morphemes belonging to the same morpheme tag obtained from the test set by using the trained CRF model are given in Table~\ref{ex_tag}. The final morpheme tags show that allomorphs can ben learned by our model. For example, \textit{la}, \textit{le}, \textit{yla}, and \textit{yle} are all allomorphs. 

\begin{table}[!t]
\centering
\caption{Some of the morphemes and their tags (with frequencies) obtained from our CRF model.}
\begin{tabular}{ c | l  }
\hline
\textbf{Morpheme tag} & \textbf{Example morphemes obtained by CRF} \\
\hline
Location:					 & da (7439), de (7633), nde (4877), te (1520), nda (8548), \\ & ta (1537), \c{c}i (2), un (1) \\
\hline
Infinitive:                   & mek (2774), mak (3481)  \\
\hline
Inst:                   & la (2218), le (2299), yla (2186), yle (2518)   \\
\hline
AfterDoing:             & y\i{}p (141), yip (104), up (335), \"{u}p (106), yup (20), \"{u}n (12), \\ & \i{}m (24), \"{u}m\"{u}z (2) \\
\hline
Dative:                    & na (5877), e (7298), ne (4860), ye (2718), ya (2741)       \\
\hline
Progressive:                   & \i{}yor (4738), iyor (6022), uyor (1756), \"{u}yor (1343), \i{}c\i{} (2)  \\
\hline
Desire:                    & se (301), sa (659), yacak (8), yecek (3)   \\
\hline
Ablative:                    & nden (1959), dan (3053), tan (1061), ndan (3514), \\ & ten (874), dik (1)  \\
\hline
Narrative:                   & m\i{}\c{s} (595), m\"{u}\c{s} (276), mi\c{s} (1640), mu\c{s} (782), t\"{u}r (7), t\i{}r (30), \\ & tik\c{c}e (3), \i{}ver (2), t\i{}k (1), \"{u}ver (1), \i{}c\i{} (2), \"{u}n (1), y\i{}ver (1), \\ & t\i{}\u{g} (1), s\i{}n (1), \c{c}i (1), d\"{u}r (1), \"{u}m (1) \\
\hline
\end{tabular}
\label{ex_tag}
\end{table}

\begin{table}[t]
\centering
\caption{Morpheme tagging F1 scores for different training set sizes on Metu-Sabanc{\i} Turkish Treebank}
\begin{tabular}{ c  c  c  c  }
\hline
Train set size & Test set size & Number of tags & F1 Score  \\
\hline
500                               & 46440                         & 84                           & 80.71\%                       \\
1000                              & 45940                         & 84                           & 83.92\%                       \\
1500                              & 45440                         & 84                           & 88.48\%                       \\
2000                              & 44940                         & 84                           & 90.39\%                       \\
3000                              & 43940                         & 84                           & 92.13\%                       \\
4000                              & 42940                         & 84                           & 93.26\%                       \\
5000                              & 41940                         & 84                           & \textbf{94.12\%}                      \\
\hline
\end{tabular}
\label{crf_results_metusabanci}
\end{table}
 
We used Zemberek again in order to create gold sets for the morpheme tagging task. F1 scores for morphological tagging for different sizes of train and test sets obtained from Metu-Sabanc{\i} Turkish Treebank are given in Table~\ref{crf_results_metusabanci}\footnote{Precision and recall values are the same because gold sets and results consists of same number of morphemes since we are only doing morpheme tagging and not any segmentation.}. The F1 score of the model is $80.7\%$ for a training set with 500 words, whereas the F1 score increases up to $94.1\%$ on a 5K training set. Therefore, the F1 score significantly improves on the larger training sets.

We also tested our model on the manually collected newspaper archives, which is much larger than Metu-Sabanc{\i} Turkish Treebank. We obtained an F1 score of $93.7\%$ on a 5K train set and 700K test set. This shows that the performance of our model does not drop significantly for larger test sets. The results are given in Table~\ref{crf_results_boun}.

\begin{table}[!t]
\centering
\caption{Results of morpheme tagging on manually collected newspaper archives.}
\begin{tabular}{ c  c  c  c  }
\hline
Train set size            & Test set size                  & Number of tags               & F1 Score                   \\
\hline
5000 & 720332 & 88 & 93.70\% \\
\hline
\end{tabular}
\label{crf_results_boun}
\end{table}

We compared our model with Chipmunk~\cite{Ryan} by using their tag set and datasets. The results are given in Table~\ref{comp_chipmunk_crf}. Our CRF model outperforms their model on accuracy, whereas their model outperforms ours on F1 score. However, it should be noted that Chipmunk dataset lacks the derivation morpheme tags, whereas we are also learning derivation morpheme tags in our model. 

\begin{table}[!t]
\centering
\caption{Comparison of Chipmunk~\cite{Ryan} and our CRF model for morpheme tagging} 
\begin{tabular}{ c  c  c c  c }
\hline
& Train set size  & Test set size & Accuracy & F1 Score \\
\hline
Chipmunk		 & 1987                              & 1586                          & 56.06\%                        & \textbf{85.07\%}              \\
CRF            & 1987                              & 1586                          & \textbf{66.62\%}                        & 66.62\%  \\
\hline   
\end{tabular}
\label{comp_chipmunk_crf}
\end{table}

\subsection{Experiments on PoS Tagging}

Our PoS tag set consists of 13 major PoS tags~\cite{ehsani}, that are Adj, Adv, Conj, Det, Interj, Noun, Num, Postp, Pron, Punc, Verb, Ques, Dup. 

\begin{table}[t]
\centering
\caption{PoS tagging accuracy scores on Metu-Sabanc{\i} Turkish Treebank}
\begin{tabular}{ c  c  c  c  c  c }
\hline
Train size  & Test size   & Tag num. & Word Emission Acc.    & Suffix  Acc.        & Morpheme Tag Acc.                   \\
\hline
5025 & 1017 & 13 &84.85\% & 86.23\% & \textbf{88.98\%} \\
18205 & 1017 & 13 &88.59\% & 88.69\% & \textbf{90.95\%} \\
39392 & 1017 & 13 &89.18\% & 89.57\% & \textbf{91.05\%} \\
\hline
\end{tabular}
\label{hmm_results_metu}
\end{table}

We tested our model on two different datasets. The first set of experiments were performed on Metu-Sabanc{\i} Turkish Treebank. The results are given in Table~\ref{hmm_results_metu} for different sizes of train/test sets and for different emission types. We provide results for word emissions, last suffix emissions, and the tag of the last morpheme's emissions. For a 5K training set, word emission accuracy is $84.8\%$, suffix emission accuracy is $86.2\%$, and morpheme tag emission accuracy is $88.9\%$. This shows that using morpheme tag emission outperforms both word emissions and last suffix emissions in smaller datasets. The accuracy increases on a $\sim$40K train set, but still using morpheme tag emissions outperforms using word and last suffix emissions. 

The results obtained from the manually collected newspaper archives are given in Table~\ref{hmm_results_boun}. This time using the word emissions outperforms using the last suffix and the last morpheme tag emissions because the sparsity becomes no longer a problem in the larger train sets.

\begin{table}[t]
\centering
\caption{PoS tagging accuracy scores on manually collected newspaper archives}
\begin{tabular}{ c  c  c  c  c   c }
\hline
Train size  & Test size   & Tag num. & Word Emission Acc.    & Suffix  Acc.        & Morpheme Tag Acc.                   \\
\hline
5677 & 1005 & 13 & 83.88\% & 86.66\% & \textbf{89.25\%} \\
25535 & 1005 & 13 &90.64\% & 89.45\% & \textbf{91.94\%} \\
53829 & 1005 & 13 &92.43\% & 91.11\% & \textbf{92.93\%} \\
106019 & 1005 & 13 &\textbf{94.62\%} & 91.64\% & 93.43\% \\
714757 & 1005 & 13 &\textbf{95.44\%} & 91.34\% & 93.83\% \\
\hline
\end{tabular}
\label{hmm_results_boun}
\end{table}

\begin{table}[t]
\centering
\caption{PoS Tagging accuracy scores for the experiments with/without terminal punctuation on Metu Sabanc{\i} Turkish Treebank.}
\begin{tabular}{c c c c }
\hline
                    & 5K & 18K & 39K \\ \hline
With terminal punc - multiple HMM 			&\textbf{88.98\%}       &\textbf{90.95\%}	    &\textbf{91.05\%}    \\ 
With terminal punc  - single HMM 
	&88.60\%		&90.86\%		&90.96\%  \\ 
Without terminal punc - multiple HMM    	&87.51\%		&89.74\%		&89.85\%      \\ 
Without terminal punc - single HMM     		&86.30\%		&88.19\%		&88.53\% \\ 

\hline
\end{tabular}
\label{hmm_results_comp_with_punc}
\end{table}

\begin{table}[t]
\centering
\caption{Comparison of our model with suffix based tagger~\cite{dincer} and the perceptron algorithm~\cite{Sak2007} on the datasets obtained from Metu Sabanc{\i} Turkish Treebank.}
\begin{tabular}{c c c c }
\hline
                    & Train Set Size & Test Set Size & Accuracy \\ \hline
Suffix-based tagger~\cite{dincer} & 5025           & 1017          & 84.25\%    \\ 
Perceptron~\cite{Sak2007}          & 5025           & 1017          & 85.15\%    \\ 
HMM with the last morpheme tag       & 5025           & 1017          & \textbf{88.98\%}    \\ 
\hline
Suffix-based tagger~\cite{dincer} & 18205           & 1017          & 88.90\%    \\ 
Perceptron~\cite{Sak2007}          & 18205           & 1017          & 86.71\%    \\ 
HMM with the last morpheme tag      & 18205           & 1017          & \textbf{90.95\%}    \\ 
\hline
\end{tabular}
\label{hmm_results_comp}
\end{table}

In order to measure the impact of terminal punctuation in PoS tagging, we did two sets of experiments on Metu Sabanc{\i} Turkish Treebank. In the first set of experiments, we included the terminal punctuation, whereas in the second set of experiments we excluded the terminal punctuation. While including the terminal punctuation, first we built one HMM for each sentence in the training set, second we built only one HMM for the entire corpus where all the words are linked to each other on the same HMM that are separated by terminal punctuation. We obtained an accuracy of $88.9\%$ for multiple HMM approach, whereas we obtained an accuracy of $88.6\%$ for a single HMM approach on 5K train set. In the second set of experiments, we completely excluded the terminal punctuation and repeated the experiments for multiple HMMs and a single HMM. We obtained an accuracy of $87.5\%$ for multiple HMMs, whereas we obtained an accuracy of $86.3\%$ for a single HMM. Results are given in Table~\ref{hmm_results_comp_with_punc}. It shows that even though terminal punctuation plays an important role in PoS tagging, behaving each sentence as a single HMM by assuming that sentences are independent from each other leads to a slight increase in the accuracy.

We compared our model with Sak et al.~\cite{Sak2007} and Dincer et al.~\cite{dincer} on Metu Sabanc{\i} Turkish Treebank. We used the last 5 letters of each word with a second order HMM to implement the suffix based tagger by Dincer et al.~\cite{dincer}, since their model gives the best scores for the last 5 letters. Results are given in Table~\ref{hmm_results_comp}. The results show that our model outperforms the other two models on smaller datasets (i.e. 5K and 18K). 

Obtaining data is one major problem in natural language processing tasks. Using small datasets by reducing sparsity is one challenge in natural language processing. Here, we aimed to increase the accuracy of PoS tagging for an agglutinating language on smaller datasets when large datasets are not available. Our results show that it is possible to use a kind of class-based language model by grouping the morphemes according to their syntactic roles within a word by tagging them and then using it for PoS tagging to reduce the sparsity in smaller datasets.

\section{Conclusion \& Future Work}
\label{conclusion}

We introduced a CRF model to tag the morphemes syntactically and a HMM model for PoS tagging that uses these morpheme tags in order to reduce the sparsity in Turkish PoS tagging on smaller datasets. We managed to obtain morpheme tags with F1 score 94.1\% on a limited training set by using CRFs. Then, we trained a second-order HMM model with the last morpheme tag of each word emitted from each HMM state in order to perform PoS tagging, contrary to the conventional approach of using words' surface forms emitted from HMM states. The results show that using the last morpheme tags helps dealing with the sparsity especially on small train sets.  

We believe that morphological features of the context words will be also informative in morpheme tagging task because Eryigit et al.~\cite{eryigit} shows that using inflectional groups as units in Turkish dependency parsing increases the parsing performance. We leave using the contextual information in morpheme tagging as a future work. 

\section{Acknowledgments}
This research is supported by the Scientific and Technological Research Council of Turkey (TUBITAK) with the project number EEEAG-115E464 and we are grateful to TUBITAK for their financial support.

%
\bibliographystyle{splncs03}
\bibliography{main}

\end{document}